\documentclass[11pt,a4paper]{article}
\usepackage[hyperref]{emnlp2018}
\usepackage{times}
\usepackage{latexsym}
\usepackage{graphicx}
\usepackage{url}
\usepackage{amsfonts}
\usepackage{amsmath}
\usepackage{caption}
\usepackage{subcaption}
\usepackage[export]{adjustbox}

\usepackage{url}

\aclfinalcopy 

\newcommand{\possessivecite}[1]{\citeauthor{#1}'s (\citeyear{#1})}

\newcommand{\ignore}[1]{}

\title{Style Transfer Through Multilingual and Feedback-Based Back-Translation}

\author{Shrimai Prabhumoye, Yulia Tsvetkov, Alan W Black, Ruslan Salakhutdinov\\
  Carnegie Mellon University, Pittsburgh, PA, USA\\
  {\tt \{sprabhum, ytsvetko, awb, rsalakhu\}@cs.cmu.edu}
  }

\date{}

\begin{document}
\maketitle
\begin{abstract}

Style transfer is the task of transferring an attribute
of a sentence (e.g., formality) while maintaining its semantic content. 
The key challenge in style transfer is to strike a balance
between the competing goals, one to preserve meaning and the other to improve the style transfer accuracy. 
Prior research has identified that the task of meaning preservation is generally harder to attain and evaluate. 
This paper proposes two extensions of the state-of-the-art 
style transfer models aiming at improving the meaning preservation in style transfer. 
Our evaluation shows that these extensions help to ground meaning better while improving the transfer accuracy. 
\end{abstract}

\section{Introduction}
Consider the following two comments about a movie: 
(1) \textit{I entered the theater in the bloom of 
youth and emerged with a family of field mice living
in my long, white mustache};\footnote{From \url{https://www.thestranger.com/movies/1210980/sex-and-the-city-2}} and 
(2) \textit{The movie was very long}. 
Although the meaning of the two sentences is 
similar,
their styles are very different. 
Style transfer is the task of transferring the attributes of
a sentence (e.g., `sarcastic' and `not-sarcastic') without changing its content. 
It is important for dialog systems such as personalized agents, customer service agents and smart home 
assistants to generate responses that are fluent and 
fit the social setting.


Advances in text generation has motivated recent work on style transfer with non-parallel corpora \cite{shen2017style,hu2017toward,FuTan,HeHe}. 
\citet{shen2017style} propose a novel method which leverages the refined alignment of latent representations to perform style transfer.
The paper introduces cross-aligned auto-encoder with discriminators. 
\citet{hu2017toward} learn a disentangled latent representation and use a code to generate a sentence.
\citet{FuTan} explore two models for style transfer which use multiple decoders or style embeddings to augment the encoded representations. 
\citet{style_transfer_acl18} propose to transfer style through back-translation. 
The latter method is simpler to train and it attains the state-of-the-art performance in style transfer accuracy, confirming the efficacy of back-translation in grounding meaning.
The goal of the current study is to investigate alternative back-translation setups that attain a better balance between meaning preservation and style transfer. 

We introduce two approaches which extend the back-translation models proposed by \citet{style_transfer_acl18} exploring back-translation setups that preserve the content of the sentence better. 
The first approach explores multilingual pivoting, hypothesizing that transfer through several languages will help ground meaning better than transfer through one language. We follow \possessivecite{johnson2017google} setup. 
The second approach is an investigation to include a term in the loss function which corresponds to preserving semantic content of the sentence: we add a \textit{feedback} loss to the generative models.
We evaluate our models along three dimensions: style transfer accuracy, fluency and preservation of meaning.
We compare the results with the cross-aligned auto-encoder 
\cite{shen2017style} and the back-translation model with one pivot language \cite{style_transfer_acl18}.
We find that both extensions improve the accuracy of style transfer without reduction in preservation of meaning.


\section{Grounding Meaning In Back-Translation}
\label{sec:multi}

\begin{figure}[t]
\centering
\includegraphics[width=0.45\textwidth]{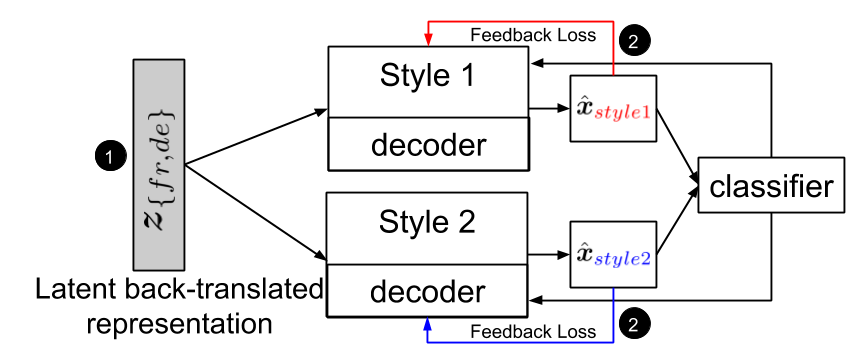}
\caption{The two extension to the back-translation model in the train pipeline. (1) Improving the latent representation through multi-lingual back-translation. Here, the latent representation is created using two pivot languages -- French and German. (2) Extending the original model to add feedback loss to the generators.}
\label{extension}
\end{figure}

While the previous work \cite{style_transfer_acl18} focuses on creating a representation by translating to a pivot language, preserving meaning in the generated sentences is still an unsolved question. 
In this work, we try to tackle this question by extending their model in two directions: 
(1) To improve the latent representation such that it grounds the meaning better and 
(2) Providing the generative models with a feedback which represents how good the generator performs in preserving the meaning. 
Both the extensions are marked in Figure \ref{extension}.

\paragraph{Notation.} Given a dataset $\boldsymbol{X}$ in which each instance is labeled with a style 
 $\boldsymbol{s}_1$ or $\boldsymbol{s}_2$, the goal of style transfer is to 
 generate sentences of the target style without changing the meaning of the original sentence. 
 Let the set of sentences in $\boldsymbol{X}$ which belong to $\boldsymbol{s}_1$ be 
 $\boldsymbol{X}_1 = \{\boldsymbol{x}_{1}^{(1)},\ldots,\boldsymbol{x}_{1}^{(n)}\}$ 
 and the sentences which belong to $\boldsymbol{s}_2$ be  
 $\boldsymbol{X}_2 = \{\boldsymbol{x}_{2}^{(1)},\ldots,\boldsymbol{x}_{2}^{(n)}\}$. 
 We denote the sentences of $\boldsymbol{X}_1$ transferred to style $\boldsymbol{s}_2$ as 
 $\hat{\boldsymbol{X}_{12}} = \{\hat{\boldsymbol{x}}_{12}^{(1)},\ldots,\hat{\boldsymbol{x}}_{12}^{(n)}\}$ 
 and the sentences of  $\boldsymbol{X}_2$ transferred to style $\boldsymbol{s}_1$  by 
 $\hat{\boldsymbol{X}_{21}} = \{\hat{\boldsymbol{x}}_{21}^{(1)},\ldots,\hat{\boldsymbol{x}}_{21}^{(n)}\}$.

\paragraph{Style transfer through Back-translation.}
\citet{style_transfer_acl18} introduces the technique of 
back-translation to perform style transfer.
They first transfer a sentence to one pivot language and 
use the encoding of the sentence in the pivot language to
train the generative models corresponding to the two styles.
They also use feedback from a pre-trained classifier to guide
the generators to generate the desired style.
The objective function of the generative models is:
\begin{equation}
\text{min}_{\theta_{gen}} \mathcal{L}_{gen} = \mathcal{L}_{recon} + \lambda_{c}\mathcal{L}_{class}
\end{equation}
This model is denoted as \textbf{B}ack-translated \textbf{S}tyle \textbf{T}ransfer (\textbf{BST}) in the future.

\paragraph{Grounding meaning with multilingual back-translation.}

\citet{johnson2017google} showed
that multi-lingual neural machine translation systems using 
one-to-many and many-to-one frameworks can perform zero-shot learning.
We want to leverage this approach to ground meaning in
style transfer using multiple pivot languages.
We have trained a one to many translation system \cite{johnson2017google} where we have a encoder-decoder network for one source language and two target languages.
We also train a many to one translation system \cite{johnson2017google} where we have a encoder-decoder network for two source languages and one target language.
We use these translation systems for training the style specific decoders following the procedure in \cite{style_transfer_acl18}.
Specifically, a sentence is first translated from English to two pivot languages. 
We create the latent representation of the sentence by encoding the sentence in both pivot languages using the many to one translation system.
\begin{eqnarray}
\boldsymbol{z}_1 &=& \text{Encoder}_{mo} (\boldsymbol{x}_{l_1}) \\
\boldsymbol{z}_2 &=& \text{Encoder}_{mo} (\boldsymbol{x}_{l_2}) \\
\boldsymbol{z} &=& (\boldsymbol{z}_1 + \boldsymbol{z}_2)/2
\end{eqnarray}
where, $\text{Encoder}_{mo}$ is the encoder of the many to one translate system, 
$\boldsymbol{x}_{l_1}$ is the sentence in pivot language 1 and 
$\boldsymbol{x}_{l_2}$ is the sentence in pivot language 2. 
The final representation is given by elementwise average of the two representations.
This model is denoted as \textbf{M}ulti-lingual \textbf{B}ack-translated \textbf{S}tyle \textbf{T}ransfer (\textbf{MBST}) in the future.

\paragraph{Grounding meaning with feedback.}
This approach adds a loss function to the generative models
which guides them to generate sentences that are closer to the original sentence and hence to preserve meaning better.
The models trained by the back-translation approach are 
fine tuned to include the feedback loss function.
To provide this feedback to the generators, 
we first generate the $\hat{\boldsymbol{X}_{12}}$ 
and $\hat{\boldsymbol{X}_{21}}$ using the generative models corresponding to style 
$\boldsymbol{s}_2$ and $\boldsymbol{s}_1$ respectively.
Data $\hat{\boldsymbol{X}_{12}}$  is now representative of style $\boldsymbol{s}_2$ and 
data $\hat{\boldsymbol{X}_{21}}$ is representative of style $\boldsymbol{s}_1$.
Hence, now we have train data that is transferred from style $\boldsymbol{s}_1$ to
$\boldsymbol{s}_2$ and transferred from style $\boldsymbol{s}_2$ to $\boldsymbol{s}_1$.

We use this data to fine-tune the models. 
We use the same back-translation procedure as described in 
Section \ref{sec:multi} to first translate this data to the two pivot languages and then create a 
latent representation of the data.
While fine-tuning the generative model for style $\boldsymbol{s}_1$, 
we transfer data $\hat{\boldsymbol{X}_{12}}$ to style $\boldsymbol{s}_1$.
Let the data generated in this process be denoted by $\hat{\boldsymbol{X}_{121}}$.
Our loss function compares the generated data $\hat{\boldsymbol{X}_{121}}$ with the 
original $\boldsymbol{s}_1$ data $\boldsymbol{X}_{1}$.
Similarly, while fine-tuning the generative model for style $\boldsymbol{s}_2$, 
we transfer data $\hat{\boldsymbol{X}_{21}}$ to style $\boldsymbol{s}_2$.
Let the data generated in this process be denoted by $\hat{\boldsymbol{X}_{212}}$.
Our loss function compares the generated data $\hat{\boldsymbol{X}_{212}}$ with the 
original $\boldsymbol{s}_2$ data $\boldsymbol{X}_{2}$.
Let $\boldsymbol{\theta}_{gen}$ denote the parameters of the generators.
The generative loss $\mathcal{L}_{gen}$ is then given by:
\begin{equation}
\text{min}_{\theta_{gen}} \mathcal{L}_{gen} = \mathcal{L}_{recon} + \lambda_{c}\mathcal{L}_{class} +
\lambda_{f}\mathcal{L}_{feed}
\end{equation}
where $\mathcal{L}_{recon}$ is the reconstruction loss, $\mathcal{L}_{class}$ is the binary cross entropy loss,  $\mathcal{L}_{feed}$ is the reconstruction loss for feedback and $\lambda_{c}, \lambda{f}$ are the balancing parameters. 
This model is denoted as \textbf{M}ulti-lingual \textbf{B}ack-translated \textbf{S}tyle \textbf{T}ransfer  + \textbf{F}eedback (\textbf{MBST+F}) in the future.

\section{Experiments}
\subsection{Style Transfer Tasks}
We use three tasks described in \cite{style_transfer_acl18}
to evaluate our models.
The three tasks correspond to: (1) gender transfer: 
we transfer the style of writing reviews of Yelp from male and female authors \cite{reddy2016obfuscating}. 
(2) political slant transfer: we transfer the style of addressing comments to the two political parties namely democratic and republican \cite{rtgender} and 
(3) sentiment modification: here we focus on only two sentiments - positive and negative. The goal is to modify the sentiment of the sentence while preserving the content.

\subsection{Baselines}

Our baseline model is a \textbf{C}ross-aligned \textbf{A}uto-\textbf{E}ncoder (\textbf{CAE}) from \cite{shen2017style}. 
We use the off-the-shelf trained model for sentiment 
modification task and we separately train this model for the gender and political slant tasks. 
We also compare our results with the back-translation model 
using only one pivot language and with no feedback loss (\textbf{BST} model).

\subsection{Evaluation Tasks}
\begin{table}[t]
\begin{center}
\begin{tabular}{ l | r | r | r }
\hline
Model & Gender & Political & Sentiment \\
\hline
CAE & 60.40 & 75.82 & 80.43 \\
BST & 57.04 & 88.01 & 87.22 \\
MBST & 58.84 & 95.61 & 91.95 \\
MBST+F & \textbf{64.07} & \textbf{97.69} & \textbf{93.31}\\
\hline
\end{tabular}
\end{center}
\caption{Style transfer accuracy for generated sentences.}
\label{accuracy_result}
\end{table}

\begin{table*}[t]
\begin{center}
\begin{tabular}{ l | r | r | r | r | r | r}
\hline
Model & Fem - Mal & Mal - Fem & Dem - Rep &
Rep - Dem & Pos - Neg & Neg - Pos \\
\hline
CAE & 31.04 & 31.15 & 25.25 & 27.15 & 31.24 & 25.75  \\
BST & 27.60 & 26.67 & 23.87 & 32.10 & 18.29 & 13.30 \\
MBST & 20.76 & 21.03 & \textbf{15.32} & \textbf{17.48} &	\textbf{10.59} & 8.14  \\
MBST+F & \textbf{19.99} & \textbf{18.78} & 16.08 & 19.23 & 11.56 & \textbf{7.41} \\
\hline
\end{tabular}
\end{center}
\caption{Perplexity of generated sentences. Here Fem=Female, Mal=Male, Dem=Democratic, Rep=Republican, Pos=Positive, Neg=Negative}
\label{perplexity}
\end{table*}

\paragraph{Style Transfer Accuracy.} 
We measure the accuracy of style transfer as described in \cite{shen2017style}. 
We have reproduced the classifiers described in \cite{style_transfer_acl18}. 
The classifier has an accuracy of 82\% for the gender-annotated corpus, 92\% accuracy 
for the political slant dataset and 93.23\% accuracy for the sentiment dataset. 
We use these classifiers to test the generated sentences for the desired style.

\paragraph{Perplexity.} To measure the fluency of the generative models automatically, we use perplexity measure. We create separate language models for each of the three tasks using only the training data. We use only ngrams up to an order of 3 to create the language model \footnote{We use the SRILM toolkit \cite{stolcke2002srilm}}.

\paragraph{Meaning Preservation.} 
We follow the procedure described in \cite{bennett2005large} to perform A/B testing. 
We reuse the instructions provided by \cite{style_transfer_acl18} for the three tasks. 
But unlike \cite{style_transfer_acl18}, we perform our evaluation on Amazon Mechanical Turk.
We annotate 200 samples per task and ask 3 unique workers to annotate each sample.
We take the majority vote as the final label.
The results in \cite{style_transfer_acl18} were reproduced for comparing the \textbf{CAE} model 
with the \textbf{BST} model. As reported by them, the \textbf{BST} model
performs better in preservation of meaning for the tasks of gender and political slant transfer. 
We present the results for the comparison between \textbf{BST} and \textbf{MBST} 
models; and the \textbf{MBST} and the \textbf{MBST+F} models.

\paragraph{Fluency.} We asked human annotators on Mechanical Turk to measure the 
fluency of the generated sentences on a scale of 1 to 4.
1 is unreadable and 4 is perfectly readable. 
We annotate 120 samples for each model and each sample is 
annotated by three unique workers.
The 120 samples of each model has an equal distribution of 
samples from the three tasks.

\subsection{Experimental Setup}

We used data from Workshop in Statistical Machine Translation 2015 (WMT15) \cite{bojar-EtAl:2015:WMT} and sequence-sequence framework \cite{sutskever2014sequence,bahdanau2015neural,karpathy2015deep} to train our translation models. 
For the \textbf{BST}, we have used French as the pivot language with approximately 5.4M parallel sentences for training.
For the \textbf{MBST}, we have used French and German as the two pivot languages
We used the French--English data along with approximately 4.5M English--German parallel sentences for training. 
A vocabulary size of 100K was used to train the all translation systems. 
We have used the same setup of hyper-parameters described in that paper to compare our models fairly. 
For the style tasks, we have used the same train, dev, test and classtrain splits as used in \cite{style_transfer_acl18}.
The English to French, German translation system achieved a BLEU score \cite{koehn2007moses} of 30.92 for French and 22.85 for German translations.
The French, German to English translation system achieved a BLEU score of 31.68 and 24.90

\section{Results}

Table \ref{accuracy_result} shows the style transfer accuracy results in percentages for the generated test sentences. 
We can see the accuracy is boosted for all three tasks by the two extensions.
The \textbf{MBST+F} model performs the best in all three tasks.

Table \ref{perplexity} shows the perplexity of the generative models for each of the three tasks.
We observe that both \textbf{MBST} and \textbf{MBST+F} models are better than the \textbf{NMT} and \textbf{CAE} models 
but there is no significant difference between the two models.

\begin{table}[h!]
\begin{center}
\begin{tabular}{ l | r | r | r }
\hline
Experiment & BST & No Pref. & MBST \\
\hline
Gender & 26.5 & \textbf{55.0} & 18.5 \\
Political slant & \textbf{51.5} & 35.0 & 13.5 \\
Sentiment & 30.0 & \textbf{36.5} & 33.5 \\
\hline
Experiment & MBST & No Pref. & MBST+F \\
\hline
Gender & 22.5 & \textbf{65.0} & 12.5 \\
Political slant & 35.0 & \textbf{47.5} & 17.5 \\
Sentiment & 32.5 & \textbf{53.0} & 14.5 \\
\hline
\end{tabular}
\end{center}
\caption{Human preference for meaning preservation \%}
\label{tab:mbst-mbstf}
\end{table}

Table \ref{tab:mbst-mbstf} shows the results for human evaluation of the models \textbf{MBST} and \textbf{MBST+F} for preservation of meaning.  
Perhaps confusingly, these results show no clear preference between the models.  
This is a positive result as it means that these extensions do not degrade the meaning, in spite of them improving the style transfer accuracy.
Although we observe that \textbf{MBST} may be slightly preferred over \textbf{MBST+F}.  

Table \ref{fluency} shows the human evaluated fluency of the four models for the three tasks. We averaged the scores over the 120 samples and 3 annotators per sample. \textbf{MBST+F} performs better than the other models in 2 out of 3 tasks and \textbf{MBST} performs the best in one task -- political slant. The over-all averaged scores for the two models \textbf{MBST} and \textbf{MBST+F} is the same 3.08, whereas it is much lower 2.79 for \textbf{BST} and 2.57 for \textbf{CAE}.
\begin{table}[h!]
\begin{center}
\begin{tabular}{ l | r | r | r }
\hline
Model & Gender & Political & Sentiment \\
\hline
CAE & 2.00 & 2.82 & 2.88 \\
BST & 2.55 & 2.83 & 2.98 \\
MBST & 2.81 & \textbf{2.96} & 3.48 \\
MBST+F & \textbf{2.83} & 2.88 & \textbf{3.55} \\
\hline
\end{tabular}
\end{center}
\caption{Fluency in generated sentences.}
\label{fluency}
\end{table}

\section{Conclusion}

We have presented in this paper two extensions of the back-translation model. 
The first extension focused on creating  latent representation which is better grounded in meaning and the second extension targeted to provide a feedback to the generator which guides it to produce sentences similar to the original sentence.
Both the extensions allow us to boost the style transfer accuracy for all the three tasks considerably, while still preserving the meaning.

\bibliography{emnlp2018}
\bibliographystyle{acl_natbib_nourl}

\appendix

%
%
%

\end{document}